%% file: main.tex
\documentclass[sigplan]{acmart}

\renewcommand\footnotetextcopyrightpermission[1]{} 
\usepackage{natbib}
\setcitestyle{authoryear,open={(},close={)}}
\begin{document}

%
\title{Clinical Named Entity Recognition using Contextualized Token Representations}
\author{Yichao Zhou, Chelsea Ju, J. Harry Caufield, Kevin Shih, Calvin Chen, Yizhou Sun, Kai-Wei Chang, }
\author{Peipei Ping, and Wei Wang}
\affiliation{University of California, Los Angeles}
\email{{yz, chelseaju, lspss89221, calvinyhchen, yzsun, kwchang, weiwang} @cs.ucla.edu} \email{{jcaufield,pping}@mednet.ucla.edu}

%

%

\begin{abstract}

The clinical named entity recognition (CNER) task seeks to locate and classify clinical terminologies into predefined categories, such as diagnostic procedure, disease disorder, severity, medication, medication dosage, and sign symptom. 
CNER facilitates the study of side-effect on medications including identification of novel phenomena and human-focused information extraction.
Existing approaches in extracting the entities of interests focus on using static word embeddings to represent each word. 
However, one word can have different interpretations that depend on the context of the sentences. Evidently, static word embeddings are insufficient to integrate the diverse interpretation of a word.
To overcome this challenge, the technique of contextualized word embedding has been introduced to better capture the semantic meaning of each word based on its context.
Two of these language models, ELMo and Flair, have been widely used in the field of Natural Language Processing to generate the contextualized word embeddings on domain-generic documents.
However, these embeddings are usually too general to capture the proximity among vocabularies of specific domains.
To facilitate various downstream applications using clinical case reports (CCRs), we pre-train two deep contextualized language models, Clinical Embeddings from Language Model (C-ELMo) and Clinical Contextual String Embeddings (C-Flair) using the clinical-related corpus from the PubMed Central.  
Explicit experiments show that our models gain dramatic improvements compared to both static word embeddings and domain-generic language models.
The pre-trained embeddings of these two models will be available soon.
 
\end{abstract}


%
\maketitle
\input{section/intro.tex}
\input{section/rel.tex}
\input{section/model.tex}
\input{section/exp.tex}

\input{section/conclu.tex}

\bibliography{ref}

\end{document}

%% file: section/intro.tex
\section{Introduction}
Clinical case reports (CCRs) are written descriptions of the unique aspects of a particular clinical case~\citep{Caban-Martinez2012}. They are intended to serve as educational aids to science and medicine, as they play an essential role in sharing clinical experiences about atypical disease phenotypes and new therapies~\citep{caufield2018reference,zhou2021create}. Unlike other types of clinical documents (e.g., electronic medical records, or EMRs), CCRs generally describe single clinical narratives at a time: these are stories of diseases as they were observed and treated, written in language requiring domain familiarity but otherwise generally interpretable. Conveniently, accessing and reading any of the more than 2 million CCRs in publication does not involve the privacy responsibilities required by EMRs and other protected health information. CCRs therefore serve as rich, plentiful examples of clinical language. 

Clinical named entity recognition (CNER) is an important text mining task in the domain of biomedical natural language processing. It aims to identify clinical entities and events from the case reports. For example, in the sentence ``CT of the maxillofacial area showed no facial bone fracture.'' ``CT of the maxillofacial area'' is a ``diagnostic procedure'' and ``facial bone fracture'' belongs to the ``disease and disorder'' category. As with documents describing experimental procedures and results---often the focus of general biomedical annotated corpora such as PubTator~\citep{Wei2013}---CCRs include a large variety of entity types and potential orders of events~\citep{caufield2018reference}. Methods to better enable biomedical and clinical NLP at scale, across numerous entity types, and with generalizable approaches across topics are necessary, as single-task or single-entity type methods provide insufficient detail for comprehensive CNER. 
Fine-grained CNER supports development of precision medicine's hope to leverage advanced computer technologies to deeply digitize, curate and understand medical records and case reports~\citep{rajkomar2018scalable,bates2014big,zhou2020clinical}.

Biomedical NER (BioNER), of which CNER is a subtask, has been a focus of intense, groundbreaking research for decades but has recently undergone a methodological shift.
Its foundational methods are largely rule-based (e.g., Text Detective~\citep{Tamames2005}), dictionary-based (e.g., BioThesaurus~\citep{Liu2006} or MetaMap~\citep{metamap}), and basic statistical approaches (e.g., the C-value / NC-value method~\citep{Frantzi2000}).
Source entities for NER are sourced from extensive knowledgebases such as UMLS~\citep{Bodenreider2004} and UniProtKB~\citep{TheUniProtConsortium2017}.
Readily applicable model-based BioNER methods, including those built upon non-contextualized word embeddings such as Word2Vec and GloVe~\citep{w2v, glove} now promise to more fully address the challenges particular to the biomedical domain: concepts may have numerous names, abbreviated forms, modifiers, and variants. 
Furthermore, biomedical and clinical text assumes readers have extensive domain knowledge. Its documents follow no single structure across sources or topics, rendering their content difficult to predict.
 
These models neither avoid time-consuming feature engineering, nor make full use of semantic and syntactic information from each token's context. 
Context can thoroughly change an individual word's meaning, e.g., an ``infarction'' in the heart is a heart attack but the same event in the brain constitutes a stroke. Context is crucial for understanding abbreviations as well: ``MR'' may represent the medical imaging technique \textit{magnetic resonance}, the heart condition \textit{mitral regurgitation}, the concept of a \textit{medical record}, or simply the honorific \textit{Mister}. 
Non-contextualized word embeddings exacerbate the challenge of understanding distinct biomedical meanings as they contain only one representation per word. 
The most frequent semantic meaning within the training corpus becomes the standard representation. 

Inspired by the recent development of contextualized token representations~\citep{elmo, bert, flair} supporting identification of how the meaning of words changes based on surrounding context, we refresh the technology of CNER to better extract clinical entities from unstructured clinical text.
The deep contextualized token representations are pre-trained with a large corpus using a language model (LM) objective. 
ELMo \citep{elmo} takes word tokens as input and pre-trains them with a bidirectional language model (biLM). 
Flair \citep{flair} proposes a pre-trained character-level language model by passing sentences as sequences of characters into a bidirectional LSTM to generate word-level embeddings. 
BERT \citep{bert} is built with bidirectional multi-layered Transformer encoders on top of the WordPiece embeddings, position embeddings, and segment embeddings. 
In this paper, we address the CNER task with contextualized embeddings (i.e., starting with ELMo and Flair), then and compare structural differences in the resulting models. 
Following recent work demonstrating impressive performance and accuracy of pre-training word representations with domain-specific documents~\citep{indomain-elmo}, 
we collected domain-specific documents all related to CCRs, roughly a thousandth of PMC documents, and pre-trained two deep language models, C-ELMo and C-Flair. In this paper, we focus on the CNER task and evaluate the two language models across three datasets. Our two pre-trained language models can support applications beyond CNER, such as clinical relation extraction or question answering. 


Our contributions are as follows: 
\begin{itemize}
    \item To the best of our knowledge, we are the first to build a framework for solving clinical natural language processing tasks using deep contextualized token representations.
    \item  We pre-train two contextualized language models, C-ELMo and C-Flair for public use\footnote{The pre-trained model can be download at \url{https://drive.google.com/drive/folders/1b8PQyzTc_HUa5NRDqI6tQXz1mFXpJbMw?usp=sharing}}. We evaluate our models on three CNER benchmark datasets, MACCROBAT2018, i2b2-2010, NCBI-disease, and achieve dramatic improvements of 10.31\%, 7.50\%, and 6.94\%, respectively.
    \item We show that pre-training a language model with a light domain-specific corpus can result in better performance in the downstream CNER application, compared with domain-generic embeddings. 
\end{itemize}


%% file: section/rel.tex
\section{Related work}
\subsection{Clinical named entity recognition}
Clinical named entity recognition (CNER) is a fundamental technique to acquire knowledge from descriptions of clinical events and disease presentations from a wide variety of document types, published case reports and sets of electronic medical records. CNER has drawn broad attention, but heavy feature engineering is intentional for traditional CNER methods 
~\citep{metamap,ctakes,demner2017metamap,de2011machine}. 
In recent years, deep learning methods have achieved significant success in CNER tasks. 
\citet{zhang2018improving} leveraged transfer learning to use existing knowledege. 
\citet{cross-type} and \citet{han2020knowledge} applied another semi-supervised learning method, multi-task learning, to obtain useful information from other datasets or structural knowledge. 
\citet{xu2018improving} improved the performance of CNER by using a global attention mechanism. 
A residue dilated convolution network helped fast and accurate recognition on Chinese clinical corpus~\citep{qiu2018fast}. 
However, these deep learning methods all depend on the token representations that are not contextualized. The failure to track different semantic and syntactic meanings of each token leads to sub-optimal learning and modeling on named entity recognition. In this work, inspired by the recent development of contextualized token representations, we explore the ensemble of contextualized language models and simple deep learning methods for CNER.

\subsection{Contextualized token representations}
Deep contextualized word representation models complex characteristics of word use and how these uses vary across linguistic contexts~\citep{elmo,bert,flair}. As a result, the representation of each token is a function of the entire input sentence, which is different from the traditional word type embeddings. \citet{elmo} leveraged a two-layer bidirectional language model (biLMs) with character convolutions to construct this function. \citet{bert} followed the idea of self-attention mechanism~\citep{vaswani2017attention} and pre-trained a deep bidirectional Transformer by jointly
conditioning on both left and right context. \citet{flair} developed contextual string embeddings by leveraging the internal states of a trained character language model. So far, these deep language models has brought massive improvement in different NLP applications including question answering, relation extraction, and sentiment classification. 

Some researchers have applied contextualized embeddings to the biomedical domain. \citet{biobert} pre-trained a BioBERT with the settings of base BERT~\citep{bert} using billions of tokens from PubMed abstracts and PMC full text articles. \citet{peng2019transfer} pre-trained a BERT-based language model using the complete PubMed abstract and MIMIC III corpus, tested with ten datasets of five tasks.
These two works improved the performance of several representative biomedical text mining tasks; however, it required a large number of computational resources and inevitably a long time to train the language model. Inspired by \citep{indomain-elmo}, we pre-trained two light-loaded language models with a much smaller domain-specific clinical dataset selected from the PMC corpus.

%% file: section/model.tex
\section{Method}
In this section, we firstly introduce the architectures of both word-level and character-level language models in Section~\ref{sec:contextual}. Then we explain our CNER model in Section~\ref{sec:ner-model}.

\label{sec:methods}

\begin{figure*}[h!]
  \centering
    \includegraphics[width=0.65\linewidth]{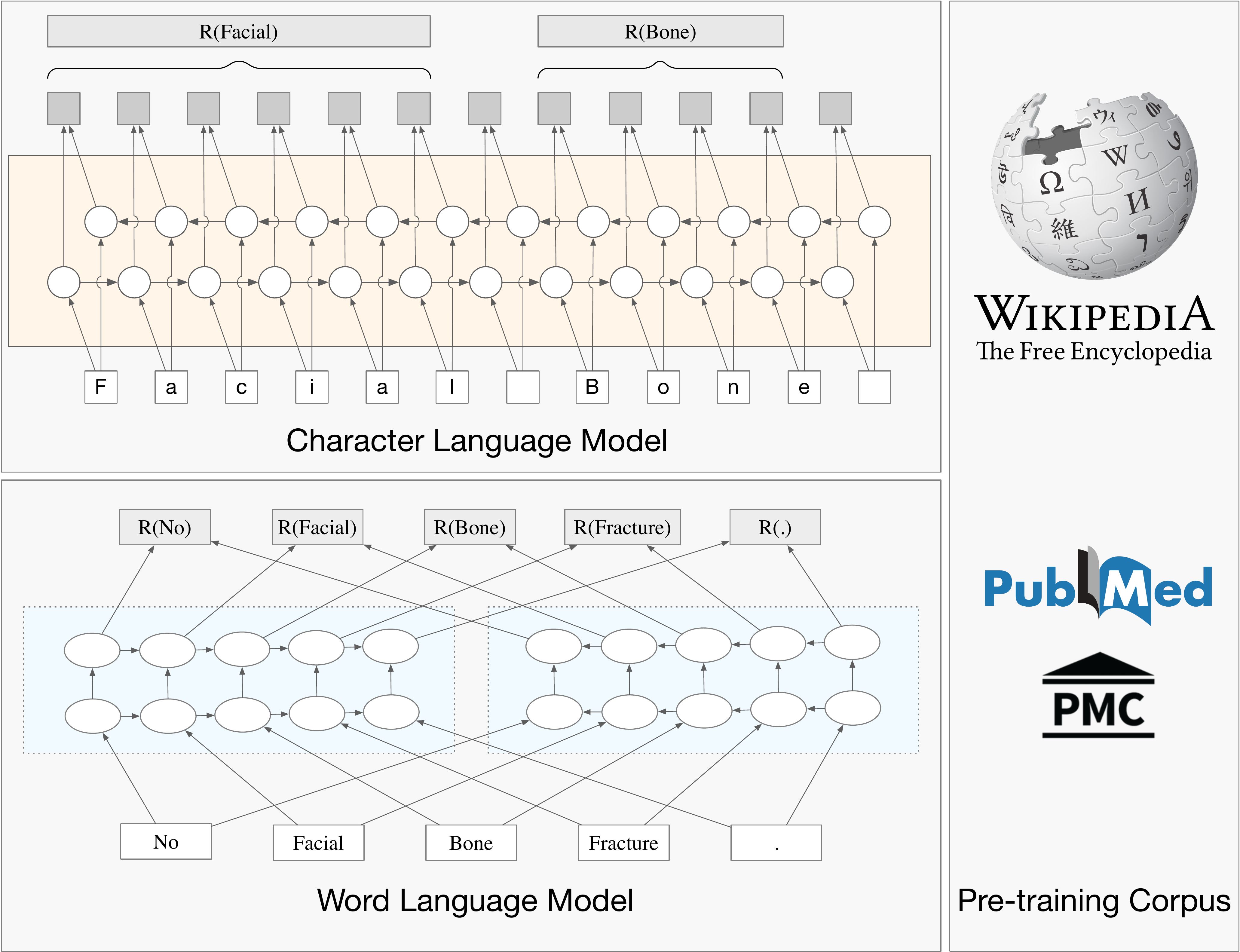}
  \caption{Character and Word Language models}
  \label{fig:LMs}
\end{figure*}

\subsection{Contextualized Embeddings}
\label{sec:contextual}
\subsubsection{ELMo}
ELMo is a language model that produces contextualized embeddings for words. It is pre-trained with a two-layered bidirectional language model (biLM) with character convolutions on a large corpus. The left lower part in Figure~\ref{fig:LMs} is the high level architecture of ELMo, where $\text{R}(\cdot)$ means the representation of a word.

ELMo takes a sequence of words $(w_1, w_2, ..., w_N)$ as input and generates context-independent token representations using a character-level CNN. Then ELMo feeds the sequence of tokens $(t_1, t_2, ..., t_N)$ into the biLM which is a bidirectional Recurrent Neural Network (RNN). The forward-LM computes the probability of each sequence by: 
\begin{equation}
    p(t_1, t_2, ..., t_N) = \prod_{k=1}^N p(t_k|t_1, t_2, ..., t_{k-1})
\end{equation} 
Thus at each position $k$, the RNN layer outputs a hidden representation $h_k$ for predicting the token $t_{k+1}$.
The backward-LM has the same structure as the forward-LM, except the input is the reverse sequence. Then, we jointly maximize the log-likelihood of both directions:
\begin{equation}
\begin{split}
    \sum_{k=1}^N &(\ \log p(t_k|t_1, t_2, ..., t_{k-1}; \theta_x, \theta_{f}, \theta_s) \\
   &+ \log p(t_k|t_{k+1}, t_{k+2}, ..., t_{N}; \theta_x, \theta_{b}, \theta_s)\ )
\end{split}
\end{equation} 
where $\theta_x$ is the token representation and $\theta_s$ is the Softmax layer for both the forward and backward LM's, and $\theta_{f}$ and $\theta_{b}$ denotes the parameters of RNNs in two  directions. 

\subsubsection{Flair}
Flair is a character-level word representation model that also uses RNN as the language modeling structure. Different from ELMo, Flair treats the text as a sequence of characters. 

The goal of most language models is to estimate a good distribution $p(t_0,t_2,...,t_T)$ where $t_0, t_1,...,t_n$ is a sequence of words. Instead of computing the distribution of words, Flair aims to estimate the probability $p(x_0,x_1,...x_T)$, where $x_0, x_1, ..., x_T$ is a sequence of characters. The joint distribution over the entire sentence can then be represented as follows:
\begin{equation}
    p(x_0, x_1, ..., x_T) = \prod_{t=0}^Tp(x_t|x_1, x_2, ..., x_{t-1})
\end{equation} 
where $p(x_t|x_0,...,x_{t-1})$ is approximated by the network output $h_t$ from one RNN layer.
\begin{equation}
    p(x_t| x_0, ..., x_{t-1}) = \prod_{t=0}^Tp(x_t|h_t;\theta)
\end{equation} 
$h_t$ is the hidden state that records the entire history of the sequence, which is computed recursively with a memory cell. $\theta$ denotes all the parameters of the RNN model. On top of the hidden layer, there is a fully-connected softmax layer, so the likelihood of a character is defined as:
\begin{equation}
    p(x_t| h_t; W)=\text{softmax}(W\cdot h_t+b)
\end{equation} 
where $W$ and $b$ are the weights and biases.

Besides, Flair also has a backward RNN layer.  Flair extracts the token embeddings from the hidden states of both the forward and backward models. Given a word that starts at index $t_s$ and ends at $t_e$ in a sequence of characters, the embeddings of this word are defined as a concatenation of the hidden states from both forward and backward models:
\begin{equation}
    r^{Flair}:=h^f_{t_e+1}\oplus h^b_{t_s-1}
\end{equation}
where $h^f$ denotes the hidden states from the forward model and $h^b$ are the hidden states from the backward model. The details are illustrated in the left upper part in Figure~\ref{fig:LMs}.

\subsection{CNER Model}
\label{sec:ner-model}
We used a well-established BiLSTM-CRF sequence tagging model \citep{bilstm-crf,cross-type,habibi2017deep} to address the downstream sequence labeling tasks. 

First, it passes sentences to a user-defined token embedding model, which converts a sequence of tokens into word embeddings: $r_0, r_1, r_2, ..., r_n$. We may concatenate embedding vectors from different sources to form a new word vector. For example, the concatenated embeddings of GloVe and Flair is represented as:
\begin{equation}
    r_i= r_i^{GloVe} \oplus r_i^{Flair} 
\end{equation}
Then, the concatenated embeddings are passed to the BiLSTM-CRF sequence labeling model to extract the entity types. 

%% file: section/exp.tex
\section{Experiments}
\label{sec:exp}
In this section, we first introduce three benchmark datasets, MACCROBAT2018, i2b2-2010 and NCBI-disease. Then, we explain the corpus we used to pre-train the language models. The performance comparison among a set of baseline models and our methods is discussed in Section~\ref{sec:results}. 
\subsection{Datasets}


\subsubsection{MACCROBAT2018}
\label{sec:MACCR}
\citet{caufield2019comprehensive} developed a standardized metadata template and identified text corresponding to medical concepts within 3,100 curated CCRs spanning 15 disease groups and more than 750 reports of rare diseases. MACCROBAT2018 is a subset of the case reports which were annotated by clinical experts. In total, there are 200 annotated case reports and 3,652 sentences containing 24 different entity/event types. We randomly selected 10\% case reports as development set and 10\% as test set. The remaining documents are used to train the CNER model. Detailed description is shown in Table~\ref{tab:table1}, \ref{tab:table2}, and \ref{tab:table3}.

\subsubsection{i2b2-2010}
\label{sec:i2b2}
The i2b2/VA 2010 Workshop \citep{i2b2-2010} on NLP challenges for clinical records presented three tasks: a concept extraction task, an assertion classification task, and a relation classification task. The i2b2-2010 dataset provides ``layered'' linguistic annotation over a set of clinical notes. In this study, we focus on the first task: given the plain text, we extract the clinical entities. The dataset contains three entity types which are ``test'', ``problem'', ``treatment''. We followed \citet{i2b2-2010} to split the dataset into train/development/test sets. 



\subsubsection{NCBI-disease}
\label{sec:ncbi}
The NCBI-disease \citep{ncbi-disease} dataset is fully annotated at the mention and concept level to serve as a research resource for the biomedical natural language processing community. The dataset contains 793 PubMed abstracts with 6,892 disease mentions which leads to 790 unique disease concepts. Therefore, the dataset only has one types which is ``disease''.

\subsection{Pre-training Corpus}
\label{sec:pretrain-data}
To pre-train the two language models, we obtained articles through the PubMed Central (PMC) FTP server\footnote{ftp://ftp.ncbi.nlm.nih.gov/pub/pmc}, and in total picked 47,990 documents that are related to clinical case reports. We indexed these documents with some keyword including ``case report'' and ``clinical report''. This corpus contains 0.1 billion words which is around 1/10 of the corpus used for the domain-generic ELMo~\citep{elmo} and Flair~\citep{flair}.
We will release our pre-trained language models soon. 

\subsection{Pre-trained Language Model}
\label{sec:pretrained-lm}
We proposed C-ELMo and C-Flair, which are respectively a pre-trained ELMo and a pre-trained Flair with the domain-specific corpus. To fairly compare the two models, we do not initialize C-ELMo and C-Flair with any pre-trained ELMo and Flair, and pre-train them on the same clinical case report corpus described in Section~\ref{sec:pretrain-data}. Moreover, we tried to set both models' parameter sizes to a similar scale. Since Flair's parameter size is 20M when it performs at its best (hidden size of 2048), we chose the medium size ELMo model correspondingly, which has 25M parameters according to AllenNLP~\cite{elmo}. All models were pre-trained on one NVIDIA Tesla V100 (16GB), with each requiring roughly one week to complete.

For C-Flair, we followed the default settings of Flair, a hidden size of 2048, a sequence length of 250, and a mini-batch size of 100. The initial learning rate is 20, and the annealing factor is 4.

For C-ELMo, we chose the medium-size model among all configurations, which has a hidden size of 2048 and projection dimension of 256. For the convolutional neural network token embeddings, the maximum length of a word is 50 and the embedding dimension is 16.

\begin{table}[h!]
    \caption{Number of sentences in each CNER dataset}
    \begin{tabular}{lccc}
        \hline
            Dataset Name & Train  & Dev   & Test  \\
        \hline
            MACCROBAT2018 & 2,894 & 380 & 351 \\ 
            i2b2-2010 &  14,683 & 1,632 & 27,626 \\ 
            NCBI-disease & 5,423 & 922 & 939 \\
        \hline
    \end{tabular}
    \label{tab:table1}
\end{table}

\begin{table}[h!]
    \caption{Number of tokens in each CNER dataset}
    \begin{tabular}{lccc}
        \hline
            Dataset Name & Train  & Dev   & Test  \\
        \hline
            MACCROBAT2018 & 64,879 & 862 & 7,955\\ 
            i2b2-2010 & 134,586 & 14,954 & 267,250 \\ 
            NCBI-disease & 135,701 & 23,969 & 24,497 \\
        \hline
    \end{tabular}

    \label{tab:table2}
\end{table}

\begin{table}[h!]
    \caption{Number of entity types in each CNER dataset}
    \begin{tabular}{lc}
        \hline
            Dataset Name & \# of Entity Types \\ 
        \hline
            MACCROBAT2018     & 24                 \\ 
            i2b2-2010         & 3                  \\ 
            NCBI-disease & 1                  \\ 
        \hline
    \end{tabular}

    \label{tab:table3}
\end{table}

\subsection{Results}
\label{sec:results}
To fairly compare the performance of each model, we pre-trained C-Flair and C-ELMo on the same subset of PubMed Central (PMC) documents. We then applied the BiLSTM-CRF model \citep{bilstm-crf} to evaluate the downstream sequence labeling tasks. The results of our experiments are shown in Table~\ref{tab:table5}. Note that ``Embeddings'' in Table~\ref{tab:table5} denotes the stacking embeddings which can be the concatenation of different word embedding vectors. 
We used the pre-trained GloVe embeddings of 100 dimensions~\footnote{\url{https://nlp.stanford.edu/projects/glove/}}. The Flair embeddings are pre-trained with a 1-billion word corpus~\citep{chelba2013one}.  
ELMo denotes the pre-trained medium-size ELMo on the same 1-billion word corpus and ELMoPubMed denotes the pre-trained ELMo model with the full PubMed and PMC corpus~\footnote{\url{https://allennlp.org/elmo/}}. We used the micro F1-score as the evaluation metric.

\begin{table*}[h!]
    \caption{The comparison of F1-scores (\%) on three datasets among different types of embeddings}
    \begin{tabular}{lccc}
        \hline
            Embeddings & MACCROBAT2018 & i2b2-2010 & NCBI-disease\\
        \hline
            GloVe      & 59.63 & 81.35 & 82.18\\
            ELMo       & 61.69 & 84.61 & 84.50\\
            Flair      & 57.25 & 81.65 & 84.23\\
            GloVe+ELMo & 63.09 & 84.82 & 85.37\\
            GloVe+Flair & 62.63 & 81.21 & 85.58\\
            GloVe+ELMoPubMed & 64.56 & 86.50 & 87.04\\ 
            \hline
            GloVe+C-ELMo & \textbf{65.75} & 87.29 & \textbf{87.88}\\
            GloVe+C-Flair & 64.18 & \textbf{87.45} & 86.60\\ 
        \hline
    \end{tabular}
    \label{tab:table5}
\end{table*}

\begin{table*}[h!]
    \caption{The performance of three baseline methods and our best model on three datasets. Our models only leverage a simple LSTM-CRF sequence labeling module with the pre-trained contextualized embeddings.}
    \begin{tabular}{lccc}
        \hline
            Models & MACCROBAT2018 & i2b2-2010 & NCBI-disease\\
        \hline
            Our best model & \textbf{65.75} & \textbf{87.45} & 87.88\\
            \citet{cnn} & 60.13 & 81.41 & 82.62 \\
            \citet{cross-type} & 63.10 & 84.97 & 86.14 \\
            \citet{biobert} & 64.38 & 86.46 & \textbf{89.36} \\
             
        \hline
    \end{tabular}
    \label{tab:table6}
\end{table*}

\subsubsection{Domain-specific  v.s. Domain-generic corpus}
From Table~\ref{tab:table5}, we can observe that the models pre-trained on the selected case report corpus outperformed all the other language models pre-trained on the domain-generic corpus. The concatenated embedding of GloVe and C-ELMo performs the best on MACCROBAT2018 and NCBI-disease datasets, while GloVe plus C-Flair achieved the best performance on i2b2-2010.  
We can conclude that pre-training the language models with a small domain-specific corpus can be more efficient and effective for improving the performance of some downstream tasks. The domain-specific knowledge can alter the distribution and the proximity among words, thus contributing a better understanding of the relationship between word and entity types in our task.

\subsubsection{Contextualized v.s. Non-contextualized embeddings}

We also used the static word embeddings, GloVe itself, to represent the tokens in the sequence labeling task. The results in \autoref{tab:table5} show that the stacking contextualized embeddings dramatically boosted the F1-score on three different datasets by 10.31\%, 7.50\%, and 6.94\%. It proves that the deep language models absorb more intensive semantic and syntactic knowledge from the contexts. We noticed that the F1-score of Flair on MACCROBAT2018 dataset was surprisingly low. It showed that the performance of a purely character-level language model may be not as robust as the word-level models.

\subsubsection{Compared with other baseline models}
\citet{cnn} proposed a bi-directional LSTM-CNNs-CRF model to make use of both word- and character-level representations. \citet{cross-type} leveraged multi-task learning and attention mechanisms to improve the performance of biomedical sequence labeling task. Compared with these two state-of-the-art models, as shown in Table~\ref{tab:table6}, our methods perform consistently better. We suppose that with the help of pre-trained contextualized embeddings, even a light-loaded downstream model can achieve extraordinary performances. 

The BioBERT proposed in \citep{biobert} was pre-trained using a language model with around 110M parameters and using a large number of computational resources (8 NVIDIA V100 32GB GPUs). However, this contextualized language model only gets better performance in the simplest dataset (NCBI-disease) with only one entity type. 
On MACCROBAT2018 and i2b2-2010, we improved the performance by 2.13\% and 1.15\%. This shows that good experimental results can be achieved by making rational use of limited resources.

\subsection{Case Study and Analysis}
We analyze the C-Flair and C-ELMo on specific categories for the MACCROBAT2018 dataset. We look into the F1-scores of 10 different entity types. All these types appear more than 50 times in the dataset.
\begin{table}[h!]
    \caption{The comparison of F1-scores (\%) between C-ELMo and C-Flair on different entity types of MACCROBAT2018}
    \begin{tabular}{lccc}
        \hline
            Entity & GloVe+C-ELMo & GloVe+C-Flair \\
        \hline
            Biological Structure & 63.94 & \textbf{64.88} \\
            Detailed Description & \textbf{45.81} & 40.00 \\ 
            Diagnostic Procedure & \textbf{74.93} & 74.71\\ 
            Disease Disorder & \textbf{50.84} & 50.83 \\ 
            Dosage      & 77.42 & \textbf{80.00} \\
            Lab Value & \textbf{74.48} & 72.31 \\
            Medication & \textbf{76.34} & 72.13 \\
            Non-biological Location & \textbf{80.77} & 76.00 \\ 
            Severity  & \textbf{72.41} & 61.81 \\
            Sign Symptom & \textbf{62.27} & 60.64 \\
        \hline
    \end{tabular}

    \label{tab:table7}
\end{table}

From Table~\ref{tab:table7}, we can see that the character-level language model C-Flair shows an advantage in the type ``Dosage''. We find that this entity type has a number of entities that do not appear in the word-level vocabulary, such as ``60 mg/m2'', ``0.5 mg'', and ``3g/d''. On the other hand, C-ELMo has a better performance in the type ``Severity'', which contains words like ``extensive'', ``complete'', ``significant'', and ``evident''. C-ELMo also extensively outperforms C-Flair in ``Detailed Description''. The representations of tokens rely more on the word-level context in these types. Therefore, C-ELMo shows better power of capturing the relationship between the word-level contextual features with the entity types.

We noticed in Table~\ref{tab:table7}, ``Disease Disorder'' achieved around 50\% F1-score with both models. Though they performed well on NCBI-disease dataset, it is hard for them to correctly recognize complex phrase-level disease entities on MACCROBAT2018, such as ``Scheuer stage 3'', and ``feeding difficulties''.

%% file: section/conclu.tex
\section{Conclusion}
In our study, we showed that contextual embeddings show a sizable advantage against non-contextual embeddings for clinical NER. In addition, pre-training a language model with a domain-specific corpus results in better performance in the downstream CNER task, compared to the off-the-shelf corpus. We also developed a comparatively fair comparison between C-ELMo and C-Flair. 

%% file: main.bbl

\begin{thebibliography}{37}


\ifx \showCODEN    \undefined \def \showCODEN     #1{\unskip}     \fi
\ifx \showDOI      \undefined \def \showDOI       #1{#1}\fi
\ifx \showISBNx    \undefined \def \showISBNx     #1{\unskip}     \fi
\ifx \showISBNxiii \undefined \def \showISBNxiii  #1{\unskip}     \fi
\ifx \showISSN     \undefined \def \showISSN      #1{\unskip}     \fi
\ifx \showLCCN     \undefined \def \showLCCN      #1{\unskip}     \fi
\ifx \shownote     \undefined \def \shownote      #1{#1}          \fi
\ifx \showarticletitle \undefined \def \showarticletitle #1{#1}   \fi
\ifx \showURL      \undefined \def \showURL       {\relax}        \fi
\providecommand\bibfield[2]{#2}
\providecommand\bibinfo[2]{#2}
\providecommand\natexlab[1]{#1}
\providecommand\showeprint[2][]{arXiv:#2}

\bibitem[\protect\citeauthoryear{Akbik, Blythe, and Vollgraf}{Akbik
  et~al\mbox{.}}{2018}]%
        {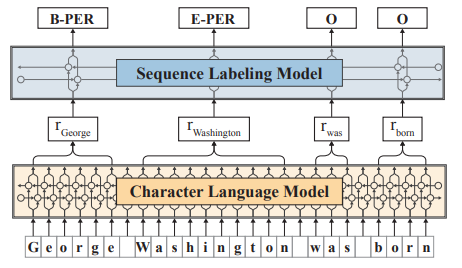}
\bibfield{author}{\bibinfo{person}{Alan Akbik}, \bibinfo{person}{Duncan
  Blythe}, {and} \bibinfo{person}{Roland Vollgraf}.}
  \bibinfo{year}{2018}\natexlab{}.
\newblock \showarticletitle{Contextual String Embeddings for Sequence
  Labeling}. In \bibinfo{booktitle}{\emph{Proceedings of the 27th International
  Conference on Computational Linguistics}}. \bibinfo{publisher}{Association
  for Computational Linguistics}, \bibinfo{pages}{1638--1649}.
\newblock


\bibitem[\protect\citeauthoryear{Aronson}{Aronson}{2001}]%
        {metamap}
\bibfield{author}{\bibinfo{person}{A.~R. Aronson}.}
  \bibinfo{year}{2001}\natexlab{}.
\newblock \showarticletitle{Effective mapping of biomedical text to the UMLS
  Metathesaurus: the MetaMap program.}
\newblock \bibinfo{journal}{\emph{Proc AMIA Symp}} (\bibinfo{year}{2001}),
  \bibinfo{pages}{17--21}.
\newblock
\showISSN{1531-605X}
\urldef\tempurl%
\url{http://view.ncbi.nlm.nih.gov/pubmed/11825149}
\showURL{%
\tempurl}


\bibitem[\protect\citeauthoryear{Bates, Saria, Ohno-Machado, Shah, and
  Escobar}{Bates et~al\mbox{.}}{2014}]%
        {bates2014big}
\bibfield{author}{\bibinfo{person}{David~W Bates}, \bibinfo{person}{Suchi
  Saria}, \bibinfo{person}{Lucila Ohno-Machado}, \bibinfo{person}{Anand Shah},
  {and} \bibinfo{person}{Gabriel Escobar}.} \bibinfo{year}{2014}\natexlab{}.
\newblock \showarticletitle{Big data in health care: using analytics to
  identify and manage high-risk and high-cost patients}.
\newblock \bibinfo{journal}{\emph{Health Affairs}} \bibinfo{volume}{33},
  \bibinfo{number}{7} (\bibinfo{year}{2014}), \bibinfo{pages}{1123--1131}.
\newblock


\bibitem[\protect\citeauthoryear{Bodenreider}{Bodenreider}{2004}]%
        {Bodenreider2004}
\bibfield{author}{\bibinfo{person}{Olivier Bodenreider}.}
  \bibinfo{year}{2004}\natexlab{}.
\newblock \showarticletitle{The Unified Medical Language System (UMLS):
  integrating biomedical terminology}.
\newblock \bibinfo{journal}{\emph{Nucleic Acids Research}}
  \bibinfo{volume}{32}, \bibinfo{number}{90001} (\bibinfo{year}{2004}),
  \bibinfo{pages}{D267--70}.
\newblock
\urldef\tempurl%
\url{https://doi.org/10.1093/nar/gkh061}
\showDOI{\tempurl}


\bibitem[\protect\citeauthoryear{Cab\'{a}n-Martinez and
  Garc\'{i}a-Beltr\'{a}n}{Cab\'{a}n-Martinez and
  Garc\'{i}a-Beltr\'{a}n}{2012}]%
        {Caban-Martinez2012}
\bibfield{author}{\bibinfo{person}{Alberto~J Cab\'{a}n-Martinez} {and}
  \bibinfo{person}{Wilfredo~F Garc\'{i}a-Beltr\'{a}n}.}
  \bibinfo{year}{2012}\natexlab{}.
\newblock \showarticletitle{Advancing medicine one research note at a time: the
  educational value in clinical case reports}.
\newblock \bibinfo{journal}{\emph{BMC Research Notes}} \bibinfo{volume}{5},
  \bibinfo{number}{1} (\bibinfo{year}{2012}), \bibinfo{pages}{293}.
\newblock
\urldef\tempurl%
\url{https://doi.org/10.1186/1756-0500-5-293}
\showDOI{\tempurl}


\bibitem[\protect\citeauthoryear{Caufield, Zhou, Bai, Liem, Garlid, Chang, Sun,
  Ping, and Wang}{Caufield et~al\mbox{.}}{2019}]%
        {caufield2019comprehensive}
\bibfield{author}{\bibinfo{person}{J~Harry Caufield}, \bibinfo{person}{Yichao
  Zhou}, \bibinfo{person}{Yunsheng Bai}, \bibinfo{person}{David~A Liem},
  \bibinfo{person}{Anders~O Garlid}, \bibinfo{person}{Kai-Wei Chang},
  \bibinfo{person}{Yizhou Sun}, \bibinfo{person}{Peipei Ping}, {and}
  \bibinfo{person}{Wei Wang}.} \bibinfo{year}{2019}\natexlab{}.
\newblock \showarticletitle{A Comprehensive Typing System for Information
  Extraction from Clinical Narratives}.
\newblock \bibinfo{journal}{\emph{medRxiv}} (\bibinfo{year}{2019}),
  \bibinfo{pages}{19009118}.
\newblock


\bibitem[\protect\citeauthoryear{Caufield, Zhou, Garlid, Setty, Liem, Cao, Lee,
  Murali, Spendlove, Wang, et~al\mbox{.}}{Caufield et~al\mbox{.}}{2018}]%
        {caufield2018reference}
\bibfield{author}{\bibinfo{person}{J~Harry Caufield}, \bibinfo{person}{Yijiang
  Zhou}, \bibinfo{person}{Anders~O Garlid}, \bibinfo{person}{Shaun~P Setty},
  \bibinfo{person}{David~A Liem}, \bibinfo{person}{Quan Cao},
  \bibinfo{person}{Jessica~M Lee}, \bibinfo{person}{Sanjana Murali},
  \bibinfo{person}{Sarah Spendlove}, \bibinfo{person}{Wei Wang},
  {et~al\mbox{.}}} \bibinfo{year}{2018}\natexlab{}.
\newblock \showarticletitle{A reference set of curated biomedical data and
  metadata from clinical case reports}.
\newblock \bibinfo{journal}{\emph{Scientific data}}  \bibinfo{volume}{5}
  (\bibinfo{year}{2018}), \bibinfo{pages}{180258}.
\newblock


\bibitem[\protect\citeauthoryear{Chelba, Mikolov, Schuster, Ge, Brants, Koehn,
  and Robinson}{Chelba et~al\mbox{.}}{2013}]%
        {chelba2013one}
\bibfield{author}{\bibinfo{person}{Ciprian Chelba}, \bibinfo{person}{Tomas
  Mikolov}, \bibinfo{person}{Mike Schuster}, \bibinfo{person}{Qi Ge},
  \bibinfo{person}{Thorsten Brants}, \bibinfo{person}{Phillipp Koehn}, {and}
  \bibinfo{person}{Tony Robinson}.} \bibinfo{year}{2013}\natexlab{}.
\newblock \showarticletitle{One billion word benchmark for measuring progress
  in statistical language modeling}.
\newblock \bibinfo{journal}{\emph{arXiv preprint arXiv:1312.3005}}
  (\bibinfo{year}{2013}).
\newblock


\bibitem[\protect\citeauthoryear{De~Bruijn, Cherry, Kiritchenko, Martin, and
  Zhu}{De~Bruijn et~al\mbox{.}}{2011}]%
        {de2011machine}
\bibfield{author}{\bibinfo{person}{Berry De~Bruijn}, \bibinfo{person}{Colin
  Cherry}, \bibinfo{person}{Svetlana Kiritchenko}, \bibinfo{person}{Joel
  Martin}, {and} \bibinfo{person}{Xiaodan Zhu}.}
  \bibinfo{year}{2011}\natexlab{}.
\newblock \showarticletitle{Machine-learned solutions for three stages of
  clinical information extraction: the state of the art at i2b2 2010}.
\newblock \bibinfo{journal}{\emph{Journal of the American Medical Informatics
  Association}} \bibinfo{volume}{18}, \bibinfo{number}{5}
  (\bibinfo{year}{2011}), \bibinfo{pages}{557--562}.
\newblock


\bibitem[\protect\citeauthoryear{Demner-Fushman, Rogers, and
  Aronson}{Demner-Fushman et~al\mbox{.}}{2017}]%
        {demner2017metamap}
\bibfield{author}{\bibinfo{person}{Dina Demner-Fushman},
  \bibinfo{person}{Willie~J Rogers}, {and} \bibinfo{person}{Alan~R Aronson}.}
  \bibinfo{year}{2017}\natexlab{}.
\newblock \showarticletitle{MetaMap Lite: an evaluation of a new Java
  implementation of MetaMap}.
\newblock \bibinfo{journal}{\emph{Journal of the American Medical Informatics
  Association}} \bibinfo{volume}{24}, \bibinfo{number}{4}
  (\bibinfo{year}{2017}), \bibinfo{pages}{841--844}.
\newblock


\bibitem[\protect\citeauthoryear{Devlin, Chang, Lee, and Toutanova}{Devlin
  et~al\mbox{.}}{2018}]%
        {bert}
\bibfield{author}{\bibinfo{person}{Jacob Devlin}, \bibinfo{person}{Ming{-}Wei
  Chang}, \bibinfo{person}{Kenton Lee}, {and} \bibinfo{person}{Kristina
  Toutanova}.} \bibinfo{year}{2018}\natexlab{}.
\newblock \showarticletitle{{BERT:} Pre-training of Deep Bidirectional
  Transformers for Language Understanding}.
\newblock \bibinfo{journal}{\emph{CoRR}}  \bibinfo{volume}{abs/1810.04805}
  (\bibinfo{year}{2018}).
\newblock


\bibitem[\protect\citeauthoryear{Do\u{g}an, Leaman, and Lu}{Do\u{g}an
  et~al\mbox{.}}{2014}]%
        {ncbi-disease}
\bibfield{author}{\bibinfo{person}{Rezarta~Islamaj Do\u{g}an},
  \bibinfo{person}{Robert Leaman}, {and} \bibinfo{person}{Zhiyong Lu}.}
  \bibinfo{year}{2014}\natexlab{}.
\newblock \showarticletitle{NCBI Disease Corpus}.
\newblock \bibinfo{journal}{\emph{J. of Biomedical Informatics}}
  \bibinfo{volume}{47}, \bibinfo{number}{C} (\bibinfo{date}{Feb.}
  \bibinfo{year}{2014}).
\newblock


\bibitem[\protect\citeauthoryear{Frantzi, Ananiadou, and Mima}{Frantzi
  et~al\mbox{.}}{2000}]%
        {Frantzi2000}
\bibfield{author}{\bibinfo{person}{Katerina Frantzi}, \bibinfo{person}{Sophia
  Ananiadou}, {and} \bibinfo{person}{Hideki Mima}.}
  \bibinfo{year}{2000}\natexlab{}.
\newblock \showarticletitle{Automatic recognition of multi-word terms:. the
  C-value/NC-value method}.
\newblock \bibinfo{journal}{\emph{International Journal on Digital Libraries}}
  \bibinfo{volume}{3}, \bibinfo{number}{2} (\bibinfo{year}{2000}),
  \bibinfo{pages}{115--130}.
\newblock
\urldef\tempurl%
\url{https://doi.org/10.1007/s007999900023}
\showDOI{\tempurl}


\bibitem[\protect\citeauthoryear{Habibi, Weber, Neves, Wiegandt, and
  Leser}{Habibi et~al\mbox{.}}{2017}]%
        {habibi2017deep}
\bibfield{author}{\bibinfo{person}{Maryam Habibi}, \bibinfo{person}{Leon
  Weber}, \bibinfo{person}{Mariana Neves}, \bibinfo{person}{David~Luis
  Wiegandt}, {and} \bibinfo{person}{Ulf Leser}.}
  \bibinfo{year}{2017}\natexlab{}.
\newblock \showarticletitle{Deep learning with word embeddings improves
  biomedical named entity recognition}.
\newblock \bibinfo{journal}{\emph{Bioinformatics}} \bibinfo{volume}{33},
  \bibinfo{number}{14} (\bibinfo{year}{2017}), \bibinfo{pages}{i37--i48}.
\newblock


\bibitem[\protect\citeauthoryear{Han, Zhou, and Peng}{Han
  et~al\mbox{.}}{2020}]%
        {han2020knowledge}
\bibfield{author}{\bibinfo{person}{Rujun Han}, \bibinfo{person}{Yichao Zhou},
  {and} \bibinfo{person}{Nanyun Peng}.} \bibinfo{year}{2020}\natexlab{}.
\newblock \showarticletitle{Domain Knowledge Empowered Structured Neural Net
  for End-to-End Event Temporal Relation Extraction}. In
  \bibinfo{booktitle}{\emph{Proceedings of the 2020 Conference on Empirical
  Methods in Natural Language Processing (EMNLP)}}.
  \bibinfo{publisher}{Association for Computational Linguistics},
  \bibinfo{address}{Online}, \bibinfo{pages}{5717--5729}.
\newblock
\urldef\tempurl%
\url{https://doi.org/10.18653/v1/2020.emnlp-main.461}
\showDOI{\tempurl}


\bibitem[\protect\citeauthoryear{Huang, Xu, and Yu}{Huang
  et~al\mbox{.}}{2015}]%
        {bilstm-crf}
\bibfield{author}{\bibinfo{person}{Zhiheng Huang}, \bibinfo{person}{Wei Xu},
  {and} \bibinfo{person}{Kai Yu}.} \bibinfo{year}{2015}\natexlab{}.
\newblock \showarticletitle{Bidirectional {LSTM-CRF} Models for Sequence
  Tagging}.
\newblock \bibinfo{journal}{\emph{CoRR}}  \bibinfo{volume}{abs/1508.01991}
  (\bibinfo{year}{2015}).
\newblock


\bibitem[\protect\citeauthoryear{Lee, Yoon, Kim, Kim, Kim, So, and Kang}{Lee
  et~al\mbox{.}}{2019}]%
        {biobert}
\bibfield{author}{\bibinfo{person}{Jinhyuk Lee}, \bibinfo{person}{Wonjin Yoon},
  \bibinfo{person}{Sungdong Kim}, \bibinfo{person}{Donghyeon Kim},
  \bibinfo{person}{Sunkyu Kim}, \bibinfo{person}{Chan~Ho So}, {and}
  \bibinfo{person}{Jaewoo Kang}.} \bibinfo{year}{2019}\natexlab{}.
\newblock \showarticletitle{BioBERT: a pre-trained biomedical language
  representation model for biomedical text mining}.
\newblock \bibinfo{journal}{\emph{CoRR}}  \bibinfo{volume}{abs/1901.08746}
  (\bibinfo{year}{2019}).
\newblock


\bibitem[\protect\citeauthoryear{Liu, Hu, Torii, Wu, and Friedman}{Liu
  et~al\mbox{.}}{2006}]%
        {Liu2006}
\bibfield{author}{\bibinfo{person}{H. Liu}, \bibinfo{person}{Z.-Z. Hu},
  \bibinfo{person}{M. Torii}, \bibinfo{person}{C. Wu}, {and}
  \bibinfo{person}{C. Friedman}.} \bibinfo{year}{2006}\natexlab{}.
\newblock \showarticletitle{Quantitative Assessment of Dictionary-based Protein
  Named Entity Tagging}.
\newblock \bibinfo{journal}{\emph{Journal of the American Medical Informatics
  Association}} \bibinfo{volume}{13}, \bibinfo{number}{5}
  (\bibinfo{year}{2006}), \bibinfo{pages}{497--507}.
\newblock
\urldef\tempurl%
\url{https://doi.org/10.1197/jamia.M2085}
\showDOI{\tempurl}


\bibitem[\protect\citeauthoryear{Ma and Hovy}{Ma and Hovy}{2016}]%
        {cnn}
\bibfield{author}{\bibinfo{person}{Xuezhe Ma} {and} \bibinfo{person}{Eduard~H.
  Hovy}.} \bibinfo{year}{2016}\natexlab{}.
\newblock \showarticletitle{End-to-end Sequence Labeling via Bi-directional
  LSTM-CNNs-CRF}.
\newblock \bibinfo{journal}{\emph{CoRR}}  \bibinfo{volume}{abs/1603.01354}
  (\bibinfo{year}{2016}).
\newblock


\bibitem[\protect\citeauthoryear{Mikolov, Sutskever, Chen, Corrado, and
  Dean}{Mikolov et~al\mbox{.}}{2013}]%
        {w2v}
\bibfield{author}{\bibinfo{person}{Tomas Mikolov}, \bibinfo{person}{Ilya
  Sutskever}, \bibinfo{person}{Kai Chen}, \bibinfo{person}{Greg~S Corrado},
  {and} \bibinfo{person}{Jeff Dean}.} \bibinfo{year}{2013}\natexlab{}.
\newblock \showarticletitle{Distributed Representations of Words and Phrases
  and their Compositionality}.
\newblock In \bibinfo{booktitle}{\emph{Advances in Neural Information
  Processing Systems 26}}, \bibfield{editor}{\bibinfo{person}{C.~J.~C. Burges},
  \bibinfo{person}{L.~Bottou}, \bibinfo{person}{M.~Welling},
  \bibinfo{person}{Z.~Ghahramani}, {and} \bibinfo{person}{K.~Q. Weinberger}}
  (Eds.). \bibinfo{pages}{3111--3119}.
\newblock


\bibitem[\protect\citeauthoryear{Peng, Yan, and Lu}{Peng et~al\mbox{.}}{2019}]%
        {peng2019transfer}
\bibfield{author}{\bibinfo{person}{Yifan Peng}, \bibinfo{person}{Shankai Yan},
  {and} \bibinfo{person}{Zhiyong Lu}.} \bibinfo{year}{2019}\natexlab{}.
\newblock \showarticletitle{Transfer Learning in Biomedical Natural Language
  Processing: An Evaluation of BERT and ELMo on Ten Benchmarking Datasets}.
\newblock \bibinfo{journal}{\emph{arXiv preprint arXiv:1906.05474}}
  (\bibinfo{year}{2019}).
\newblock


\bibitem[\protect\citeauthoryear{Pennington, Socher, and Manning}{Pennington
  et~al\mbox{.}}{2014}]%
        {glove}
\bibfield{author}{\bibinfo{person}{Jeffrey Pennington},
  \bibinfo{person}{Richard Socher}, {and} \bibinfo{person}{Christopher~D.
  Manning}.} \bibinfo{year}{2014}\natexlab{}.
\newblock \showarticletitle{Glove: Global vectors for word representation}. In
  \bibinfo{booktitle}{\emph{In EMNLP}}.
\newblock


\bibitem[\protect\citeauthoryear{Peters, Neumann, Iyyer, Gardner, Clark, Lee,
  and Zettlemoyer}{Peters et~al\mbox{.}}{2018}]%
        {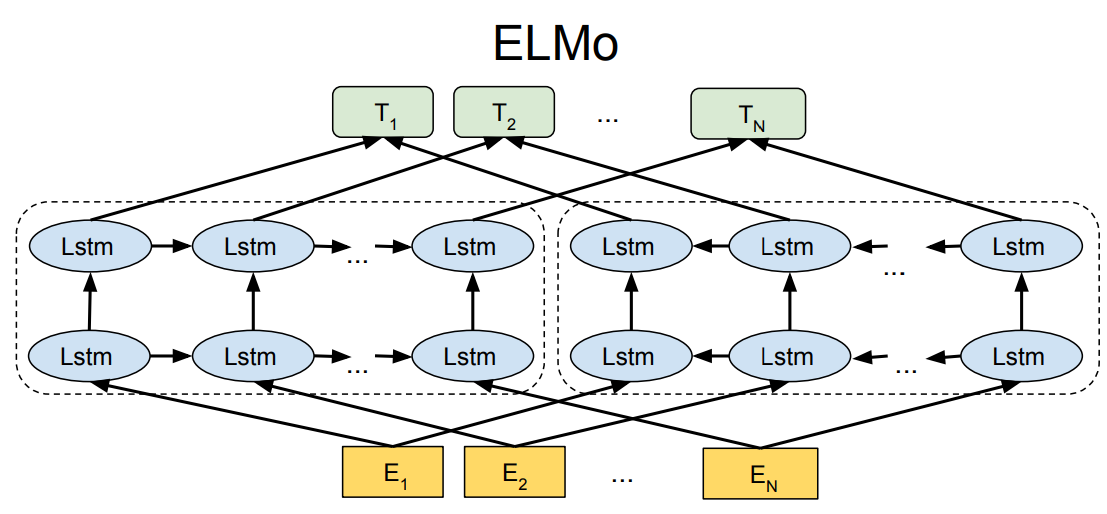}
\bibfield{author}{\bibinfo{person}{Matthew~E. Peters}, \bibinfo{person}{Mark
  Neumann}, \bibinfo{person}{Mohit Iyyer}, \bibinfo{person}{Matt Gardner},
  \bibinfo{person}{Christopher Clark}, \bibinfo{person}{Kenton Lee}, {and}
  \bibinfo{person}{Luke Zettlemoyer}.} \bibinfo{year}{2018}\natexlab{}.
\newblock \showarticletitle{Deep contextualized word representations}.
\newblock \bibinfo{journal}{\emph{CoRR}}  \bibinfo{volume}{abs/1802.05365}
  (\bibinfo{year}{2018}).
\newblock


\bibitem[\protect\citeauthoryear{Qiu, Wang, Zhou, Ruan, and Gao}{Qiu
  et~al\mbox{.}}{2018}]%
        {qiu2018fast}
\bibfield{author}{\bibinfo{person}{Jiahui Qiu}, \bibinfo{person}{Qi Wang},
  \bibinfo{person}{Yangming Zhou}, \bibinfo{person}{Tong Ruan}, {and}
  \bibinfo{person}{Ju Gao}.} \bibinfo{year}{2018}\natexlab{}.
\newblock \showarticletitle{Fast and Accurate Recognition of Chinese Clinical
  Named Entities with Residual Dilated Convolutions}. In
  \bibinfo{booktitle}{\emph{2018 IEEE International Conference on
  Bioinformatics and Biomedicine (BIBM)}}. IEEE, \bibinfo{pages}{935--942}.
\newblock


\bibitem[\protect\citeauthoryear{Rajkomar, Oren, Chen, Dai, Hajaj, Hardt, Liu,
  Liu, Marcus, Sun, et~al\mbox{.}}{Rajkomar et~al\mbox{.}}{2018}]%
        {rajkomar2018scalable}
\bibfield{author}{\bibinfo{person}{Alvin Rajkomar}, \bibinfo{person}{Eyal
  Oren}, \bibinfo{person}{Kai Chen}, \bibinfo{person}{Andrew~M Dai},
  \bibinfo{person}{Nissan Hajaj}, \bibinfo{person}{Michaela Hardt},
  \bibinfo{person}{Peter~J Liu}, \bibinfo{person}{Xiaobing Liu},
  \bibinfo{person}{Jake Marcus}, \bibinfo{person}{Mimi Sun}, {et~al\mbox{.}}}
  \bibinfo{year}{2018}\natexlab{}.
\newblock \showarticletitle{Scalable and accurate deep learning with electronic
  health records}.
\newblock \bibinfo{journal}{\emph{NPJ Digital Medicine}} \bibinfo{volume}{1},
  \bibinfo{number}{1} (\bibinfo{year}{2018}), \bibinfo{pages}{18}.
\newblock


\bibitem[\protect\citeauthoryear{Savova, Masanz, Ogren, Zheng, Sohn,
  Kipper-Schuler, and Chute}{Savova et~al\mbox{.}}{2010}]%
        {ctakes}
\bibfield{author}{\bibinfo{person}{Guergana~K Savova}, \bibinfo{person}{James~J
  Masanz}, \bibinfo{person}{Philip~V Ogren}, \bibinfo{person}{Jiaping Zheng},
  \bibinfo{person}{Sunghwan Sohn}, \bibinfo{person}{Karin~C Kipper-Schuler},
  {and} \bibinfo{person}{Christopher~G Chute}.}
  \bibinfo{year}{2010}\natexlab{}.
\newblock \showarticletitle{Mayo clinical Text Analysis and Knowledge
  Extraction System (cTAKES): architecture, component evaluation and
  applications}.
\newblock   \bibinfo{volume}{17} (\bibinfo{year}{2010}).
\newblock


\bibitem[\protect\citeauthoryear{Sheikhshabbafghi, Birol, and
  Sarkar}{Sheikhshabbafghi et~al\mbox{.}}{2018}]%
        {indomain-elmo}
\bibfield{author}{\bibinfo{person}{Golnar Sheikhshabbafghi},
  \bibinfo{person}{Inanc Birol}, {and} \bibinfo{person}{Anoop Sarkar}.}
  \bibinfo{year}{2018}\natexlab{}.
\newblock \showarticletitle{In-domain Context-aware Token Embeddings Improve
  Biomedical Named Entity Recognition}. In
  \bibinfo{booktitle}{\emph{Proceedings of the Ninth International Workshop on
  Health Text Mining and Information Analysis}}.
\newblock


\bibitem[\protect\citeauthoryear{Tamames}{Tamames}{2005}]%
        {Tamames2005}
\bibfield{author}{\bibinfo{person}{Javier Tamames}.}
  \bibinfo{year}{2005}\natexlab{}.
\newblock \showarticletitle{Text Detective: a rule-based system for gene
  annotation in biomedical texts}.
\newblock \bibinfo{journal}{\emph{BMC Bioinformatics}} \bibinfo{volume}{6},
  \bibinfo{number}{Suppl 1} (\bibinfo{year}{2005}), \bibinfo{pages}{S10}.
\newblock
\urldef\tempurl%
\url{https://doi.org/10.1186/1471-2105-6-S1-S10}
\showDOI{\tempurl}


\bibitem[\protect\citeauthoryear{{The UniProt Consortium}}{{The UniProt
  Consortium}}{2017}]%
        {TheUniProtConsortium2017}
\bibfield{author}{\bibinfo{person}{{The UniProt Consortium}}.}
  \bibinfo{year}{2017}\natexlab{}.
\newblock \showarticletitle{UniProt: the universal protein knowledgebase}.
\newblock \bibinfo{journal}{\emph{Nucleic Acids Research}}
  \bibinfo{volume}{45}, \bibinfo{number}{D1} (\bibinfo{year}{2017}),
  \bibinfo{pages}{D158--D169}.
\newblock
\urldef\tempurl%
\url{https://doi.org/10.1093/nar/gkw1099}
\showDOI{\tempurl}


\bibitem[\protect\citeauthoryear{Uzuner, South, Shen, and DuVall}{Uzuner
  et~al\mbox{.}}{2011}]%
        {i2b2-2010}
\bibfield{author}{\bibinfo{person}{{\"O}. Uzuner}, \bibinfo{person}{B.R.
  South}, \bibinfo{person}{S. Shen}, {and} \bibinfo{person}{S.L. DuVall}.}
  \bibinfo{year}{2011}\natexlab{}.
\newblock \showarticletitle{2010 i2b2/VA challenge on concepts, assertions, and
  relations in clinical text}.
\newblock \bibinfo{journal}{\emph{Journal of the American Medical Informatics
  Association}} \bibinfo{volume}{18}, \bibinfo{number}{5}
  (\bibinfo{year}{2011}), \bibinfo{pages}{552--556}.
\newblock


\bibitem[\protect\citeauthoryear{Vaswani, Shazeer, Parmar, Uszkoreit, Jones,
  Gomez, Kaiser, and Polosukhin}{Vaswani et~al\mbox{.}}{2017}]%
        {vaswani2017attention}
\bibfield{author}{\bibinfo{person}{Ashish Vaswani}, \bibinfo{person}{Noam
  Shazeer}, \bibinfo{person}{Niki Parmar}, \bibinfo{person}{Jakob Uszkoreit},
  \bibinfo{person}{Llion Jones}, \bibinfo{person}{Aidan~N Gomez},
  \bibinfo{person}{{\L}ukasz Kaiser}, {and} \bibinfo{person}{Illia
  Polosukhin}.} \bibinfo{year}{2017}\natexlab{}.
\newblock \showarticletitle{Attention is all you need}. In
  \bibinfo{booktitle}{\emph{Advances in neural information processing
  systems}}. \bibinfo{pages}{5998--6008}.
\newblock


\bibitem[\protect\citeauthoryear{Wang, Zhang, Ren, Zhang, Zitnik, Shang,
  Langlotz, and Han}{Wang et~al\mbox{.}}{2018}]%
        {cross-type}
\bibfield{author}{\bibinfo{person}{Xuan Wang}, \bibinfo{person}{Yu Zhang},
  \bibinfo{person}{Xiang Ren}, \bibinfo{person}{Yuhao Zhang},
  \bibinfo{person}{Marinka Zitnik}, \bibinfo{person}{Jingbo Shang},
  \bibinfo{person}{Curtis Langlotz}, {and} \bibinfo{person}{Jiawei Han}.}
  \bibinfo{year}{2018}\natexlab{}.
\newblock \showarticletitle{Cross-type Biomedical Named Entity Recognition with
  Deep Multi-Task Learning}.
\newblock \bibinfo{journal}{\emph{CoRR}}  \bibinfo{volume}{abs/1801.09851}
  (\bibinfo{year}{2018}).
\newblock
\showeprint[arxiv]{1801.09851}
\urldef\tempurl%
\url{http://arxiv.org/abs/1801.09851}
\showURL{%
\tempurl}


\bibitem[\protect\citeauthoryear{Wei, Kao, and Lu}{Wei et~al\mbox{.}}{2013}]%
        {Wei2013}
\bibfield{author}{\bibinfo{person}{Chih-Hsuan Wei}, \bibinfo{person}{Hung-Yu
  Kao}, {and} \bibinfo{person}{Zhiyong Lu}.} \bibinfo{year}{2013}\natexlab{}.
\newblock \showarticletitle{PubTator: a web-based text mining tool for
  assisting biocuration}.
\newblock \bibinfo{journal}{\emph{Nucleic Acids Research}}
  \bibinfo{volume}{41}, \bibinfo{number}{W1} (\bibinfo{year}{2013}),
  \bibinfo{pages}{W518--W522}.
\newblock
\urldef\tempurl%
\url{https://doi.org/10.1093/nar/gkt441}
\showDOI{\tempurl}


\bibitem[\protect\citeauthoryear{Xu, Wang, and He}{Xu et~al\mbox{.}}{2018}]%
        {xu2018improving}
\bibfield{author}{\bibinfo{person}{Guohai Xu}, \bibinfo{person}{Chengyu Wang},
  {and} \bibinfo{person}{Xiaofeng He}.} \bibinfo{year}{2018}\natexlab{}.
\newblock \showarticletitle{Improving clinical named entity recognition with
  global neural attention}. In \bibinfo{booktitle}{\emph{Asia-Pacific Web
  (APWeb) and Web-Age Information Management (WAIM) Joint International
  Conference on Web and Big Data}}. Springer, \bibinfo{pages}{264--279}.
\newblock


\bibitem[\protect\citeauthoryear{Zhang, Thurier, and Boyle}{Zhang
  et~al\mbox{.}}{2018}]%
        {zhang2018improving}
\bibfield{author}{\bibinfo{person}{Edmond Zhang}, \bibinfo{person}{Quentin
  Thurier}, {and} \bibinfo{person}{Luke Boyle}.}
  \bibinfo{year}{2018}\natexlab{}.
\newblock \showarticletitle{Improving Clinical Named-Entity Recognition with
  Transfer Learning.}
\newblock \bibinfo{journal}{\emph{Studies in health technology and
  informatics}}  \bibinfo{volume}{252} (\bibinfo{year}{2018}),
  \bibinfo{pages}{182--187}.
\newblock


\bibitem[\protect\citeauthoryear{Zhou, Chen, Zhang, Lee, Caufield, Chang, Sun,
  Ping, and Wang}{Zhou et~al\mbox{.}}{2021a}]%
        {zhou2021create}
\bibfield{author}{\bibinfo{person}{Yichao Zhou}, \bibinfo{person}{Wei-Ting
  Chen}, \bibinfo{person}{Bowen Zhang}, \bibinfo{person}{David Lee},
  \bibinfo{person}{J~Harry Caufield}, \bibinfo{person}{Kai-Wei Chang},
  \bibinfo{person}{Yizhou Sun}, \bibinfo{person}{Peipei Ping}, {and}
  \bibinfo{person}{Wei Wang}.} \bibinfo{year}{2021}\natexlab{a}.
\newblock \showarticletitle{CREATe: Clinical Report Extraction and Annotation
  Technology}.
\newblock \bibinfo{journal}{\emph{arXiv preprint arXiv:2103.00562}}
  (\bibinfo{year}{2021}).
\newblock


\bibitem[\protect\citeauthoryear{Zhou, Yan, Han, Caufield, Chang, Sun, Ping,
  and Wang}{Zhou et~al\mbox{.}}{2021b}]%
        {zhou2020clinical}
\bibfield{author}{\bibinfo{person}{Yichao Zhou}, \bibinfo{person}{Yu Yan},
  \bibinfo{person}{Rujun Han}, \bibinfo{person}{J.~Harry Caufield},
  \bibinfo{person}{Kai-Wei Chang}, \bibinfo{person}{Yizhou Sun},
  \bibinfo{person}{Peipei Ping}, {and} \bibinfo{person}{Wei Wang}.}
  \bibinfo{year}{2021}\natexlab{b}.
\newblock \showarticletitle{Clinical Temporal Relation Extraction with
  Probabilistic Soft Logic Regularization and Global Inference}.
\newblock \bibinfo{journal}{\emph{Proceedings of the AAAI Conference on
  Artificial Intelligence}} \bibinfo{volume}{35}, \bibinfo{number}{16}
  (\bibinfo{date}{May} \bibinfo{year}{2021}), \bibinfo{pages}{14647--14655}.
\newblock
\urldef\tempurl%
\url{https://ojs.aaai.org/index.php/AAAI/article/view/17721}
\showURL{%
\tempurl}


\end{thebibliography}
